\documentclass{article} 
\usepackage[final]{colm2025_conference}

\usepackage{microtype}
\usepackage{hyperref}
\usepackage{url}
\usepackage{booktabs}

\usepackage{lineno}

\usepackage{natbib}
\usepackage{hyperref}       
\usepackage{url}            
\usepackage{booktabs}       
\usepackage{amsfonts}       
\usepackage{nicefrac}       
\usepackage{microtype}      
\usepackage{xcolor}         
\usepackage{todonotes}
\usepackage{bm}
\usepackage{upgreek}
\usepackage{graphicx}
\usepackage{multirow}
\usepackage{booktabs}
\usepackage{tcolorbox}
\usepackage{amsmath,amsthm,amssymb}
\usepackage[ruled]{algorithm2e}

\newcommand{\modelname}{ILA}
\newcommand{\parameter}{\bm{\uptheta}}

\newcommand{\rmgamma}{\bm{\gamma}}
\newcommand{\ratio}{25\% }

\newcommand{\llama}{\textsc{Llama~2}}

\newtheorem{assumption}{Assumption}[section]

\newtheorem{theorem}{Theorem}[section]
\newtheorem{definition}{Definition}

\usepackage{booktabs}
\usepackage{multirow}
\usepackage{xcolor}
\usepackage{colortbl}
\usepackage{authblk}
\usepackage{enumitem}
\usepackage{marvosym}
\usepackage{caption}

\DeclareMathOperator*{\argmin}{arg\,min}

\definecolor{darkblue}{rgb}{0, 0, 0.5}
\hypersetup{colorlinks=true, citecolor=darkblue, linkcolor=darkblue, urlcolor=darkblue}

\title{Understanding Layer Significance in LLM Alignment}

\author{
    \makebox[\textwidth][c]{Guangyuan Shi\textsuperscript{1}, Zexin Lu\textsuperscript{1}, Xiaoyu Dong\textsuperscript{1}, Wenlong Zhang\textsuperscript{1}, Xuanyu Zhang\textsuperscript{2},}\\
    \makebox[\textwidth][c]{Yujie Feng\textsuperscript{1}, Xiao-Ming Wu\textsuperscript{1}\textsuperscript{\Letter}} \\
    \textsuperscript{1}Department of Computing, The Hong Kong Polytechnic University, \\
    Hong Kong S.A.R., China\\
    \textsuperscript{2}Du Xiaoman Financial, China \\
    \texttt{\{guang-yuan.shi, zexin.lu, xiaoyu.dong\}@connect.polyu.hk}, \\
    \texttt{\{wenlong.zhang, yujie.feng\}@connect.polyu.hk}, \\
    \texttt{xyz@mail.bnu.edu.cn}, ~
    \texttt{xiao-ming.wu@polyu.edu.hk}
}

\begin{document}

\ifcolmsubmission
\linenumbers
\fi

\maketitle

\begin{abstract}
Aligning large language models (LLMs) through supervised fine-tuning is essential for tailoring them to specific applications. Recent studies suggest that alignment primarily adjusts a model's presentation style rather than its foundational knowledge, indicating that only certain components of the model are significantly impacted. To uncover how alignment affects model behavior at a granular level, we propose identifying which layers within LLMs are most critical to the alignment process. Our approach, named ILA, involves learning a binary mask for the parameter changes in each layer during alignment, as an indicator of layer significance. Experimental results reveal that, despite substantial differences in alignment datasets, the important layers of a model identified by ILA exhibit nearly 90\% overlap, highlighting fundamental patterns in LLM alignment. The results also indicate that freezing non-essential layers improves overall model performance, while selectively tuning the most critical layers significantly enhances fine-tuning efficiency with minimal performance loss. Finally, we discuss how these findings extend from LLM alignment to reasoning.
\end{abstract}


\section{Introduction}

{Aligning large language models (LLMs) with specific requirements} is essential for enhancing their utility across diverse applications~\citep{luo2023wizardmath,yu2023metamath,luo2023wizardcoder,li2023starcoder,liu2024multimodal,liu2022boosting,feng2023towards}. 
Fine-tuning LLMs during the alignment process can significantly improve the models’ capabilities to meet targeted needs~\citep{bubeck2023sparks}. 
Typically, alignment involves fine-tuning the model on diverse datasets, which may include both human-curated~\citep{no_robots} and LLM-generated~\citep{alpaca} data, using approaches like instruction tuning~\citep{weifinetuned} and preference learning~\citep{bai2022training,rafailov2024direct}. Given the significant cost associated with full parameter fine-tuning, parameter-efficient fine-tuning (PEFT)~\citep{hu2021lora,chen2022revisiting,pan2024lisa} methods have emerged as a popular alternative, offering a balance between performance and resource efficiency.



Understanding what LLMs actually learn during the alignment process is crucial. \citet{zhou2023lima} posits that the majority of knowledge and capabilities are developed during the pre-training phase, with alignment primarily serving to refine the model's conversational style and formatting. Using a well-selected set of 1,000 training examples for supervised fine-tuning (SFT), they successfully produced a high-quality aligned model. Similarly, \citet{lin2023unlocking} investigated the token distribution of LLMs before and after alignment and found that most changes were related to ``stylistic tokens'', such as discourse markers and transition words, while the knowledge-intensive content largely remained untouched, coming from the base pre-trained model. These findings imply that the alignment process mainly adjusts the model's presentation style rather than modifying its foundational knowledge.

\begin{figure}[!ht]
    \centering
    \includegraphics[width=0.68\linewidth]{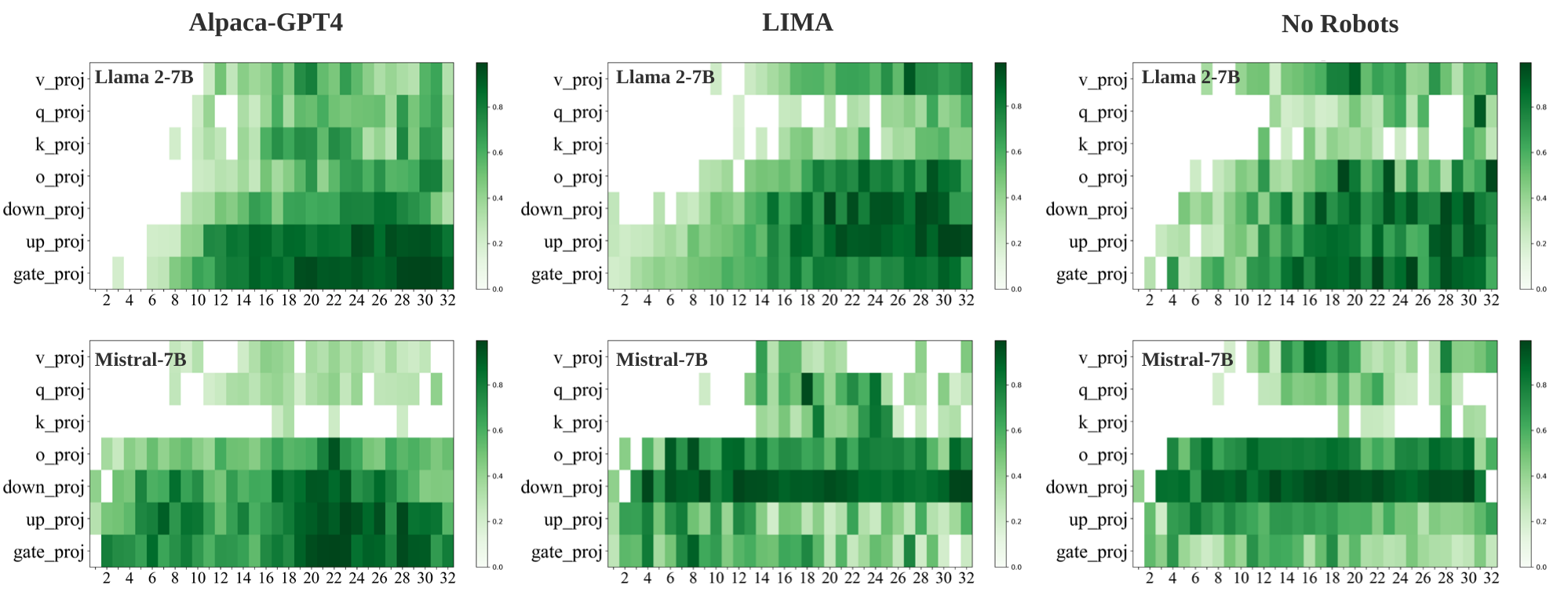}
\caption{Layer importance rankings by our \modelname{} algorithm for \textsc{Llama~\small2}-7B and Mistral-7B-v0.1 across Alpaca-GPT4, LIMA, and No Robots datasets. \textbf{Top 75\% layers by score ($s_i$) are considered important}. X-axis: transformer block index; y-axis: linear layer names. The figure highlights two findings: (1) \textbf{High overlap (90\%)} in important layers across datasets (Table\ref{table:diff_dataset}) suggests shared alignment needs, regardless of substantial differences in dataset content; (2) Important layers \textbf{differ by architecture}, reflecting model-specific dynamics.}
    \label{fig:Import_layers_datasets}
\end{figure}

\begin{table}[!t]
\renewcommand\arraystretch{0.9}
\caption{Impact of fine-tuning different components of \llama{}-7B on alignment performance using the LIMA dataset. Evaluated on MMLU (5-shot) and GPT-4o scores for Vicuna and MT-Bench prompts. Tuned components include attention projections ($W_q$, $W_k$, $W_v$, $W_o$) and feed-forward layers ($W_{\text{up}}$, $W_{\text{down}}$, $W_{\text{gate}}$).}
\label{table:Pilot_lora}
\centering
\resizebox{0.8\linewidth}{!}{
\begin{tabular}{lcccc} 
\toprule
\textbf{}                    & \begin{tabular}[c]{@{}c@{}}\textbf{ATT}\\\textbf{($W_q$, $W_k$, $W_v$, $W_o$)}\end{tabular} & \begin{tabular}[c]{@{}c@{}}\textbf{ATT2}\\\textbf{($W_q$, $W_k$, $W_v$)}\end{tabular} & \begin{tabular}[c]{@{}c@{}}\textbf{FFN}\\\textbf{($W_\text{up}$, $W_\text{down}$, $W_\text{gate}$)}\end{tabular} & \begin{tabular}[c]{@{}c@{}}\textbf{ALL}\\\textbf{(LoRA)}\end{tabular}  \\ 
\midrule
\textbf{MMLU $\uparrow$}     & 42.03                                                                                       & 42.65                                                                                 & 43.06                                                                                                            & \textbf{43.18}                                                         \\
\textbf{Vicuna $\uparrow$}   & 5.21                                                                                        & 5.13                                                                                  & 5.40                                                                                                             & \textbf{5.43}                                                          \\
\textbf{MT-Bench $\uparrow$} & 3.31                                                                                        & 3.35                                                                                  & 3.41                                                                                                             & \textbf{3.45}                                                          \\
\bottomrule
\end{tabular}
}
\end{table}

To gain a deeper understanding of LLM alignment, we analyze this process at the level of model parameters. We conducted a pilot study to investigate the impact of different model components on alignment performance, by fine-tuning only specific layers and evaluating the resulting performance, as presented in Table~\ref{table:Pilot_lora}. The results clearly indicate that fine-tuning different components of the LLM leads to considerable performance differences. For instance, fine-tuning the feed-forward network (FFN) layers achieves performance similar to fine-tuning all linear layers (i.e., with LoRA), whereas focusing solely on the attention layers causes a notable drop in performance. This observation shows the complexity of layer-specific contributions to LLM alignment, highlighting the need for detailed analysis.



To address this, we propose \textbf{\textit{identifying the layers that are most critical to alignment performance during the SFT process}}. We develop a novel approach, \modelname{}, for identifying the important layers for LLM alignment. Specifically, we learn a binary mask for the parameter changes in each layer during the fine-tuning process,
which serves as an indicator of layer significance. A binary mask value of zero indicates that the corresponding layer has negligible influence during the process, while a value of one denotes that the layer is crucial. We use gradient descent to learn the binary mask effectively and offer a theoretical analysis of the optimization process. The main findings and significance of this work include:

\begin{itemize}[leftmargin=*]
\item \textbf{Consistent layer importance ranking across different alignment datasets.} We observe similar rankings of important layers during alignment for the same pre-trained model (see Fig.~\ref{fig:Import_layers_datasets}), even though the alignment datasets vary significantly in both content and size. This suggests that the alignment process endows the model with similar capabilities, corroborating previous research findings and offers new insights into LLM alignment.

\item \textbf{Enhancing performance by freezing unimportant layers.} We show that freezing about 25\% of unimportant layers can improve performance and that \emph{a single search} for layer importance ranking is sufficient for different alignment tasks using the same architecture.

\item \textbf{Improving alignment efficiency through selective fine-tuning.} Our findings show that fine-tuning only 10-30\% key layers achieves performance comparable to fine-tuning all linear layers. Additionally, integrating this approach with QLoRA allows tuning only 30-75\% of key layers to maintain or enhance performance while cutting resource costs.

\item \textbf{Broader implications beyond LLM alignment.} Although our primary focus is on LLM alignment, the approaches and insights from this study have broader applicability. Our preliminary experiments on LLM reasoning reveal findings similar to those in alignment, showcasing the significant potential of our methods to enhance the reasoning capabilities of LLMs, particularly in achieving test-time scaling~\citep{openai2024learning,welleck2024decoding,snell2024scaling,muennighoff2025s1}.

\end{itemize}

\section{Related Works}

\textbf{LLM Alignment.}
Pretrained language models encode general-purpose representations, enabling transfer across diverse tasks~\citep{qiu2024spectral,jiang2024real,nijkamp2022codegen}. Alignment methods like instruction tuning~\citep{zhang2023instruction,sun2023comparative,muennighoff2023octopack} and preference learning~\citep{hejna2023contrastive,guan2022relative,rafailov2024direct,song2024preference,li2024more} adapt these models to specific objectives. Recent studies have explored alignment mechanisms. LIMA~\citep{zhou2023lima} showed that fine-tuning on small datasets (e.g., 1,000 examples) shapes behavior without adding new knowledge, a finding echoed by others~\citep{chen2023alpagasus,lee2023platypus,gudibande2023false}. \citet{duan2023exploring} connected instruction tuning to in-context learning via hidden state analysis, while URIAL~\citep{lin2023unlocking} revealed that alignment mainly modifies stylistic tokens, preserving knowledge-centric ones. These insights suggest alignment imparts narrow, targeted adjustments. Our work builds on this by identifying the specific layers most critical for alignment, offering a more fine-grained understanding of how adaptation occurs.

\textbf{Parameter Efficient Fine-Tuning (PEFT).}
Fine-tuning large language models with billions or trillions of parameters is computationally expensive~\citep{brown2020language,fedus2022switch}. Parameter-efficient fine-tuning (PEFT) methods address this by updating specific components~\citep{zaken2021bitfit,zhao2020masking,ansell2021composable,guo2020parameter} or using soft prompts~\citep{lester2021power,li2021prefix,asai2022attempt}. Techniques such as BitFit~\citep{zaken2021bitfit}, Adapters~\citep{houlsby2019parameter}, LoRA~\citep{hu2021lora}, and their variants~\citep{zhang2023adalora,meng2024pissa} reduce cost while maintaining transferability. Recent work~\citep{li2024gradient,hui2024hft,pan2024lisa,xu2024random,panda2024lottery} shows that selectively fine-tuning certain regions yields strong results, though random masking often lacks consistency. However, most PEFT approaches overlook parameter importance and lack prioritization. Our method addresses this by ranking layer importance, enabling targeted fine-tuning to improve performance with minimal cost.

\textbf{Layer Analysis in Model Compression.} 
Model compression techniques use structured pruning~\citep{xia2022structured,liu2024accelerating,van2023llm} and layer analysis to improve efficiency. Methods like Sheared LLaMA~\citep{xia2023sheared} and LLM-Streamline~\citep{chen2024compressing} show that pruning layers, heads, and dimensions can significantly reduce model size with minimal performance loss. Layer importance studies~\citep{zhang2024investigating,gromov2024unreasonable} further support the removal of less critical components for scalability. However, these efforts focus on reducing model size rather than optimizing parameter updates for task-specific alignment. In contrast, our work targets alignment fine-tuning by prioritizing parameter updates through skill localization~\citep{panigrahi2023task,voita2023neurons}, improving both alignment efficiency and robustness.

\section{Quantifying Layer Significance in LLM Alignment
}




To understand layer significance in LLM alignment, we propose \modelname, a method to identify important layers by learning a binary mask that indicates each layer's significance.


Consider a pre-trained LLM model with parameters $\parameter_0$ composed of $N$ layers, i.e.,
$\parameter_0=\{\parameter_0^i\}_{i=1}^{N}$. The model is fine-tuned on an alignment dataset $\mathcal{D}=\{z_i\}_{i=1}^n$ with a loss function $\mathcal{L}(\parameter)$. 
After $t$ training iterations, the model parameters are updated to $\parameter_{t} = \parameter_0+\Delta\parameter_t$, where $\Delta\parameter_t$ represents the change in parameters till iteration $t$. Define a binary mask $\rmgamma_t=\{\gamma_t^i|\gamma_t^i\in\{0,1\}\}_{i=1}^{N}$ that encodes layer-wise importance information. We apply $\rmgamma_t$ to $\Delta\parameter_t$ and define 
\begin{equation}\label{eq:mask_normal}
    \parameter^{\mathrm{mask}}_{t} = \parameter_0 + \rmgamma_t \odot \Delta\parameter_t,
\end{equation}
where $\odot$ is component-wise multiplication. The binary mask is applied to retain the changes in crucial layers while eliminating the rest. Below we provide a formal definition of the conditions under which training attains stability after an adequate number of iterations.
\begin{definition}[$\epsilon$-stable]\label{def:stable state}
$\forall \epsilon > 0$, the model is said to be $\epsilon$-stable at iteration $T$ if, for any $t > T$, the loss function satisfies the condition
\begin{equation}
    \left| \mathbb{E}_{z}[\mathcal{L}(z;\parameter_{t+1})] - \mathbb{E}_{z}[\mathcal{L}(z;\parameter_{t})] \right| < \epsilon, 
\end{equation}
where $\mathbb{E}_{z}[\cdot]$ denotes the expectation with respect to the alignment dataset $\mathcal{D}$.
\end{definition}
Once training stabilizes, we can identify the layers that are crucial for the alignment task.

\begin{definition}[Layer Importance]\label{def:skilled layers}
The binary mask $\rmgamma_t$ is defined as the solution to the following optimization problem:
\begin{equation}\label{eq:optimization}
    \rmgamma_t = \argmin_{\rmgamma_t} \mathbb{E}_{z}[\mathcal{L}(z;\parameter^\mathrm{mask}_t)], \:s.t.\:\|\rmgamma_t\|< H,
\end{equation}
where $H$ is a hyper-parameter that serves as a constraint to limit the number of important layers.
\end{definition}

\SetKwInput{KwInput}{Input} 
\SetKwInput{KwOutput}{Output}
\SetKwInput{KwInitialize}{Initialize}    

\begin{algorithm}[!t]
\renewcommand\arraystretch{0.8}
\caption{Identify the Important Layers for Alignment (\modelname)} \label{alg:MTL}
\SetAlgoLined
\KwInput{Pre-trained model parameters $\parameter_0$, learning rate $\alpha$, 
the initial importance score vector $\bm{s}_0=\{s^i_0\}_{i=1}^N$,
the number of insignificant layers $K$,
the low-rank matrices $A_0, B_0$ for the LoRA algorithm. (FFT is a special case of LoRA with full rank)
}
\BlankLine
\For{{\upshape iteration} i = {\upshape 1, 2, \dots}}{
    Update $A_{t} = A_{t-1} - \alpha \nabla_{A_{t-1}}\mathcal{L}(\parameter_t)$, $B_{t} = B_{t-1} - \alpha \nabla_{B_{t-1}}\mathcal{L}(\parameter_t)$ (LoRA)\;
    Or Update $\parameter_t=\parameter_{t-1}-\alpha\nabla_{\parameter_{t-1}}\mathcal{L}(\parameter_{t-1})$ (FFT)\;
    \If{
    Training has become stable
    }{
        Solve the optimization problem in Eq.~(\ref{eq:optim_final}) by gradient descent to find $\bm{s}_t=\{s^i_t\}_{i=1}^N$\;
        Stop training\;
    }
}
\end{algorithm}

\textbf{Efficiently Identifying the Importance Layers (Alg.~\ref{alg:MTL}).} Due to the high cost of fine-tuning large models, to address the optimization problem in Eq.~(\ref{eq:optimization}), we employ the LoRA~\citep{hu2021lora} algorithm, which utilizes low-rank decomposition matrices to represent the change in model parameters till iteration $t$ ($\Delta\parameter_t$).
Specifically, LoRA utilizes two trainable low-rank matrices, $\bm{B}^i_t\in\mathbb{R}^{d_i\times r_i}$ and $\bm{A}^i_t\in\mathbb{R}^{r_i\times k_i}$, to estimate the change of the $i^{\mathrm{th}}$ layer:
\begin{equation} \label{eq:LoRA}
    \Delta\parameter_t^{i}=\beta\cdot \bm{B}_t^i\bm{A}_t^i,
\end{equation}
where $\beta$ is the scalar hyperparameter of LoRA. With the binary mask $\rmgamma_t$, the $i^{\mathrm{th}}$ layer is updated by
\begin{equation}\label{eq:masked_lora}
    \parameter_{t}^i=\parameter_0^i + \beta\cdot \gamma^i_t \cdot \bm{B}_{t}^i\bm{A}_{t}^i.
\end{equation}
To ease the optimization of $\rmgamma_t$, we re-parametrize each of its each components $\gamma^i_t$ as the output of a Sigmoid function, i.e., $\gamma^i_t = \sigma(s^i_t)$. Then, the update of the $i^{\rm th}$ layer becomes
\begin{equation}
    \parameter_{t}^i=\parameter_0^i + \beta\cdot \sigma(s^i_t) \cdot \bm{B}_{t}^i\bm{A}_{t}^i. 
\end{equation}
Let $\bm{s}_t=\{s^i_t\}_{i=1}^{N}$, $\parameter^\mathrm{M}_t=\{\parameter_{t}^i\}_{i=1}^N$. The optimization problem in Eq.~(\ref{eq:optimization}) becomes
\begin{equation}\label{eq:optim_final}
    \bm{s}_t = \argmin_{\bm{s}_t} \mathbb{E}_{z}[\mathcal{L}(z;\parameter^\mathrm{M}_t)].
\end{equation}
We use gradient descent to optimize $\bm{s}_t$, yielding $s^i_t$ as the importance score of the $i^{\mathrm{th}}$ layer. 
A larger value of $s^i_t$ indicates $\gamma^i_t$ is closer to one, signifying higher importance of the $i^{\rm th}$ layer. 

\begin{assumption}[Lipschitz-continuous]\label{assump:L-smooth}
The loss function $\mathcal{L}(\parameter): \mathbb{R}^d\to\mathbb{R}$ is continuously differentiable and L-smooth with constant $L_1>0$ such that 
\begin{equation}
    \|\mathcal{L}(\parameter) - \mathcal{L}(\parameter')\|_2 \leq L_1\|\parameter-\parameter'\|.
\end{equation}
In addition, $\mathcal{L}(\parameter)$ has an L-Lipschitz continuous gradient with constant $L_2>0$ such that
\begin{equation}
    \|\nabla \mathcal{L}(\parameter) - \nabla \mathcal{L}(\parameter')\|_2 \leq L_2\|\parameter-\parameter'\|.
\end{equation}
\end{assumption}

\begin{assumption}\label{assump:bound}
For any $t > T$, $\parameter_{t}$ is $\epsilon$-stable. We assume there is a constant $R$ such that
\begin{equation}
    \|\parameter_t - \parameter_{t+1}\|_2 \leq R\epsilon,
\end{equation}
and there is a constant $Q$ such that $\|\parameter_t\|_2\leq Q$ for any $t>T$.
\end{assumption}

\begin{theorem}\label{theorem}
For a sufficiently small $\epsilon$, $\parameter_T$ is $\epsilon$-stable, thus Assumption~\ref{assump:L-smooth} and Assumption~\ref{assump:bound} are satisfied. For any $t>T$, we assume that $\forall i, \gamma_t^i\in[0,1]$. Let $\rmgamma_t'$ denote the result of $\rmgamma_t$ after one step of gradient descent, i.e., $\rmgamma_t'=\rmgamma_t- \beta \nabla_{\rmgamma_t}\mathcal{L}(\parameter_t^{\rm mask})$. Then we have
\begin{equation}
    \|\rmgamma_t'-\rmgamma_{t+1}'\|_2 \leq \beta (QL_2 + L_1)R\epsilon.
\end{equation}
\end{theorem}

\begin{figure*}[!t]
    \centering
    \includegraphics[width=1.0\linewidth]{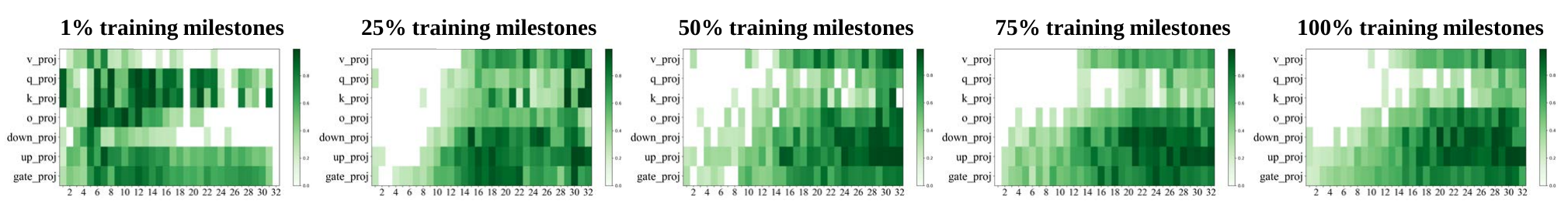}
    \caption{Layer importance rankings of \textsc{Llama~\small2}-7B during fine-tuning on LIMA at 1\%, 25\%, 50\%, 75\%, and 100\% milestones. X-axis: transformer block index; y-axis: linear layer names. Jaccard similarities are provided in Table~\ref{table:diff_stage}.}
    \label{fig:import_layers_ckpt}
\end{figure*}

This theorem demonstrates that when \(\parameter_T\) is \(\epsilon\)-stable, solving the optimization problem in Eq.~(\ref{eq:optimization}) for any \(t > T\) yields similar results. This is because, after one step of gradient descent, the difference between \(\gamma_t\) and \(\gamma_{t+1}\) is smaller than a sufficiently small number. The proof is provided in Appendix~\ref{App:Proof}, and empirical results supporting this are shown in Fig.~\ref{fig:import_layers_ckpt}.

\textbf{Leveraging Layer Importance Rankings.} The identified rankings of layer importance can be leveraged to enhanc both the performance and efficiency of LLM alignment. To maximize performance, prioritize fine-tuning the significant layers while freezing those deemed less important. For efficiency, focus on the layers most critical to model success. Detailed experiments and analyses are presented in Sec.~\ref{sec:exp}.

\section{Experiments and Findings}\label{sec:exp}

\subsection{Experimental Setup}

\textbf{Datasets.} We use three alignment datasets—Alpaca-GPT4~\citep{peng2023instruction}, LIMA~\citep{zhou2023lima}, and No Robots~\citep{no_robots}. See Appendix~\ref{App:Exp_Setup} for details.

\textbf{Models and Baselines.} We use \llama-7B/13B~\citep{touvron2023llama}, Llama 3.1-8B~\citep{dubey2024llama}, and Mistral-7B-v0.1~\citep{jiang2023mistral}. Baselines include LoRA~\citep{hu2021lora}, AdaLoRA~\citep{zhang2023adaptive}, and full fine-tuning. See Appendix~\ref{App:Exp_Setup} for details.

\textbf{Evaluation.} We evaluate (1) language understanding using MMLU~\citep{hendrycks2021measuring} and Hellaswag~\citep{zellers2019hellaswag}, and (2) conversational ability using MT-Bench~\citep{zheng2023judging} and Vicuna~\citep{chiang2023vicuna}, with scoring conducted by \textbf{GPT-4o}. Further details are provided in Appendix~\ref{App:Exp_Setup}.

\subsection{Layer Importance Rankings in LLM Alignment
} \label{subsec:exp_stability}

In this subsection, we applied \modelname{} to rank important layers during alignment across three datasets—No Robots, LIMA, and Alpaca-GPT4 (Fig.\ref{fig:Import_layers_datasets}). We also analyzed layer importance rankings at different training milestones (Fig.\ref{fig:import_layers_ckpt}). To quantify similarity between sets of important layers, we used the Jaccard similarity coefficient, defining the top 75\% highest-scoring layers as the important set $\bm{S}$. The similarity between two sets, $\bm{S}_1$ and $\bm{S}_2$, is given by:  $J(\bm{S}_1, \bm{S}_2) = \frac{|\bm{S}_1 \cap \bm{S}_2|}{|\bm{S}_1 \cup \bm{S}_2|}$. where $J=1$ indicates identical sets, and $J=0$ indicates no overlap.


\begin{table}[!t]
\renewcommand\arraystretch{0.8}
\caption{ Jaccard similarities of the top 75\% highest-scoring layers identified as important during fine-tuning of \textsc{Llama~\small2}-7B and Mistral-7B on various datasets.
} \label{table:diff_dataset}
\centering
\resizebox{0.8\linewidth}{!}{
\begin{tabular}{@{}ccccccc@{}}
\toprule
\multirow{2}{*}{Datasets} & \multicolumn{3}{c}{\textsc{Llama~\small2}-7B} & \multicolumn{3}{c}{Mistral-7B} \\ \cmidrule(l){2-7} 
                          & LIMA      & No Robots      & Alpaca-GPT4      & LIMA & No Robots & Alpaca-GPT4 \\ \midrule
LIMA                      & -         & -              & -                & -    & -         & -           \\
No Robots                 & 0.91      & -              & -                & 0.90 & -         & -           \\
Alpaca-GPT4               & 0.90      & 0.90           & -                & 0.89 & 0.93      & -           \\ \bottomrule
\end{tabular}
}
\end{table}

\textbf{Consistency in Layer Importance Rankings Across Different Datasets.} Our findings show strong consistency in layer importance rankings: (1) highly similar important layers are identified across different alignment datasets, (Fig.\ref{fig:Import_layers_datasets}, Table\ref{table:diff_dataset}); (2) the rankings remain stable across different random seeds for $\gamma$ (Table~\ref{table:diff_seed}); 
and (3) similar layers can be identified even at the beginning stages of training, such as completion of 25\% (Fig.\ref{fig:import_layers_ckpt}, Table\ref{table:diff_stage}).

These results confirm the robustness of \modelname{}, which consistently identifies stable and overlapping layers across datasets. This aligns with recent findings that alignment largely involves stylistic token shifts~\citep{lin2023unlocking}. In essence, alignment seeks similar capabilities, as \emph{evidenced by our observation that important layers remain stable across different datasets}. This underscores the relevance of our algorithm to the fundamental objectives of alignment.

\begin{table}[!t]
\renewcommand\arraystretch{0.9}
\begin{minipage}{0.35\textwidth} 
    \centering
    \caption{Jaccard similarities of the top 75\% important layers in \textsc{Llama~\small2}-7B fine-tuned on the LIMA dataset using different random seeds.}
    \label{table:diff_seed}
    \resizebox{\linewidth}{!}{
        \begin{tabular}{@{}cccc@{}}
        \toprule
        Random Seed & seed1 & seed2 & seed3 \\ \midrule
        seed1       & -     & -     & -     \\
        seed2       & 0.92  & -     & -     \\
        seed3       & 0.91  & 0.91  & -     \\ \bottomrule
        \end{tabular}
    }
\end{minipage}
\hfill 
\begin{minipage}{0.6\textwidth} 
    \centering
    \caption{Jaccard similarities of the top 75\% important layers identified at different stages of \textsc{Llama~\small2}-7B fine-tuning on the LIMA dataset.}
    \label{table:diff_stage}
    \resizebox{0.65\linewidth}{!}{
        \begin{tabular}{@{}cccccc@{}}
        \toprule
        \begin{tabular}[c]{@{}c@{}}Training \\ Milestones\end{tabular} & 1\%  & 25\% & 50\% & 75\% & 100\% \\ \midrule
        1\%                                                            & -    & -    & -    & -    & -     \\
        25\%                                                           & 0.69 & -    & -    & -    & -     \\
        50\%                                                           & 0.70 & 0.91 & -    & -    & -     \\
        75\%                                                           & 0.69 & 0.90 & 0.92 & -    & -     \\
        100\%                                                          & 0.69 & 0.91 & 0.92 & 0.93 & -     \\ \bottomrule
        \end{tabular}
    }
\end{minipage}
\end{table}

\subsection{Enhancing Alignment Performance through Freezing Unimportant Layers}

To leverage layer importance rankings, we excluded less important layers that could negatively impact fine-tuning, removing approximately \ratio of unimportant layers. The main results on \textbf{No Robots} are in Table~\ref{table:norobots_all}, with additional results for \textsc{Llama~\small2}-13B (see Table~\ref{table:llama13B}) and main results on Alpaca-GPT4 (see Table~\ref{table:alpaca_all}), and LIMA (see Table~\ref{table:lima_all}) datasets in Appendix~\ref{SubApp:Extra_Exp}. Key observations include:



(1) \textbf{Freezing unimportant layers can enhance performance}. \modelname{} consistently outperformed LoRA and full fine-tuning on most metrics, with freezing 25\% of unimportant layers yielding better results than tuning all layers. (2) \textbf{A single search for layer importance ranking suffices for a given architecture.} Rankings were stable across alignment tasks, allowing us to compute it on the No Robots dataset and apply it to others.

These results show that \modelname{} improves fine-tuning efficiency by focusing on significant layers. Compared to AdaLoRA, even though we explored a narrow range of the hyperparameter $t_r$ (target average rank of incremental matrices), our method performed better, suggesting that adjusting LoRA's matrix rank alone doesn't guarantee superior results, as also noted in~\citep{dettmers2023qlora}.

\begin{table}[!t]
\renewcommand\arraystretch{1.0}
\caption{Comparison of \textsc{Llama~\small2}-7B, Mistral-7B-v0.1, and Llama 3.1-8B fine-tuned on the No Robots dataset, evaluated on \textbf{MMLU} (5-shot), \textbf{Hellaswag} (0-shot), and \textbf{GPT-4o scores} for \textbf{Vicuna} and \textbf{MT-Bench} prompts. Vicuna and MT-Bench results are averaged over \textbf{three runs}. Grey cells indicate improvements over the base model; best scores are in bold.} \label{table:norobots_all}
\centering
\resizebox{0.95\linewidth}{!}{
\begin{tabular}{llcccc} 
\toprule
\multirow{2}{*}{\textbf{Models}} & \multirow{2}{*}{\textbf{Methods}} & \multicolumn{2}{c}{\textbf{Language Understanding}}                                                                                                                                    & \multicolumn{2}{c}{\textbf{Conversational Ability }}                                                   \\ 
\cmidrule(l){3-6}
                                 &                                   & \multicolumn{1}{l}{\textbf{MMLU $\uparrow$}}       & \multicolumn{1}{l}{\textbf{Hellaswag $\uparrow$}}                                                                                 & \textbf{Vicuna $\uparrow$}                        & \textbf{MT-Bench $\uparrow$}                       \\ 
\midrule
\multirow{5}{*}{\llama-7B}        & AdaLoRA                           & 45.23                                              & 57.30                                                                                                                             & 5.81                                              & 4.01                                               \\ 
\cmidrule(lr){2-6}
                                 & Full Fine-tune                     & 45.72                                              & 57.69                                                                                                                             & 6.12                                              & 4.18                                               \\
                                 & Full Fine-tune w/ ILA              & {\cellcolor[rgb]{0.753,0.753,0.753}}\textbf{45.98} & {\cellcolor[rgb]{0.753,0.753,0.753}}57.87                                                                                         & {\cellcolor[rgb]{0.753,0.753,0.753}}\textbf{6.35} & {\cellcolor[rgb]{0.753,0.753,0.753}}\textbf{4.37}  \\ 
\cmidrule(lr){2-6}
                                 & LoRA                              & 44.58                                              & 59.46                                                                                                                             & 5.78                                              & 4.02                                               \\
                                 & LoRA w/ ILA                       & {\cellcolor[rgb]{0.753,0.753,0.753}}45.78          & {\cellcolor[rgb]{0.753,0.753,0.753}}\textbf{59.65}                                                                                & {\cellcolor[rgb]{0.753,0.753,0.753}}5.90          & {\cellcolor[rgb]{0.753,0.753,0.753}}4.33           \\ 
\midrule
\multirow{5}{*}{Mistral-7B-v0.1} & AdaLoRA                           & 62.13                                              & 61.68                                                                                                                             & 6.21                                              & 4.69                                               \\ 
\cmidrule(lr){2-6}
                                 & Full Fine-tune                     & 61.05                                              & \textbf{64.26}                                                                                                                    & 6.32                                              & 4.55                                               \\
                                 & Full Fine-tune w/ ILA              & {\cellcolor[rgb]{0.753,0.753,0.753}}61.75          & 64.21                                                                                                                             & {\cellcolor[rgb]{0.753,0.753,0.753}}\textbf{6.51} & {\cellcolor[rgb]{0.753,0.753,0.753}}4.78           \\ 
\cmidrule(lr){2-6}
                                 & LoRA                              & 61.95                                              & 62.90                                                                                                                             & 6.25                                              & 4.68                                               \\
                                 & LoRA w/ ILA                       & {\cellcolor[rgb]{0.753,0.753,0.753}}\textbf{62.14} & {\cellcolor[rgb]{0.753,0.753,0.753}}62.98                                                                                         & {\cellcolor[rgb]{0.753,0.753,0.753}}6.42          & {\cellcolor[rgb]{0.753,0.753,0.753}}\textbf{4.87}  \\ 
\hline
\multirow{5}{*}{Llama 3.1-8B}    & AdaLoRA                           & 64.85                                              & 62.85                                                                                                                             & 6.51                                              & 5.08                                               \\ 
\cline{2-6}
                                 & Full Fine-tune                     & 64.44                                              & 63.65                                                                                                                             & 6.50                                              & 5.11                                               \\
                                 & Full Fine-tune w/ ILA              & {\cellcolor[rgb]{0.753,0.753,0.753}}65.00          & {\cellcolor[rgb]{0.753,0.753,0.753}}\begin{tabular}[c]{@{}>{\cellcolor[rgb]{0.753,0.753,0.753}}c@{}}\textbf{63.69}\\\end{tabular} & {\cellcolor[rgb]{0.753,0.753,0.753}}\textbf{6.61} & {\cellcolor[rgb]{0.753,0.753,0.753}}\textbf{5.23}  \\ 
\cline{2-6}
                                 & LoRA                              & 64.95                                              & 60.77                                                                                                                             & 6.33                                              & 4.58                                               \\
                                 & LoRA w/ ILA                       & {\cellcolor[rgb]{0.753,0.753,0.753}}\textbf{65.43} & {\cellcolor[rgb]{0.753,0.753,0.753}}60.95                                                                                         & {\cellcolor[rgb]{0.753,0.753,0.753}}6.45          & {\cellcolor[rgb]{0.753,0.753,0.753}}4.69           \\
\bottomrule
\end{tabular}
}
\end{table}


Additionally, as discussed in Sec.~\ref{subsec:exp_stability}, the stability of the layer importance ranking across datasets means a single search is often sufficient. In our experiments, we computed the layer importance ranking using full training iterations on the No Robots dataset, and then directly applied this ranking to other datasets. Though dataset-specific rankings can further improve results (Table~\ref{table:cross} in Sec.~\ref{table:efficiency_performance}), the strong cross-dataset performance with one ranking demonstrates our approach’s robustness and generalizability.

\subsection{Enhancing Alignment Efficiency by Fine-tuning Only the Most Critical Layers
}

To investigate this issue, we fine-tune the top 10\%, 20\%, and 30\% of the important layers of Mistral-7B-v0.1, as identified by \modelname{}, on the No Robots dataset, and compare the results with those of the LoRA algorithm. The results demonstrate clear benefits in focusing on a subset of important layers:


\textbf{(1) Fine-tuning a small subset of the most important layers achieves competitive performance and enhances efficiency.}
Fine-tuning the top 10\%, 20\%, or 30\% of layers results in only a slight performance drop compared to full fine-tuning. Fine-tuning 30\% of layers nearly matches full fine-tuning (Table~\ref{table:efficiency_performance}), demonstrating that focusing on the most important layers ensures efficient fine-tuning with minimal performance loss.

\textbf{(2) Our method can be applied to enhance QLoRA, further reducing costs.} When combined with QLoRA, our method fine-tunes only 30\% or 75\% of the most important layers while maintaining or improving performance (Table~\ref{table:qlora}), highlighting the efficiency of our approach in achieving comparable or better results with fewer layers.

\begin{table}
\renewcommand\arraystretch{0.8}
\caption{Fine-tuning results of Mistral-7B-v0.1 on the No Robots dataset, evaluated on MMLU (5-shot), Hellaswag (0-shot), and GPT-4o scores for Vicuna and MT-Bench prompts (averaged over three runs). Percentages in parentheses denote the fraction of fine-tuned linear layers. Best results are in bold.} \label{table:efficiency_performance}
\centering
\resizebox{0.95\linewidth}{!}{
\begin{tabular}{llcccc} 
\toprule
\multirow{2}{*}{\textbf{Models}} & \multirow{2}{*}{\textbf{Methods}} & \multicolumn{2}{c}{\textbf{\textbf{\textbf{\textbf{\textbf{\textbf{\textbf{\textbf{Language Understanding}}}}}}}}} & \multicolumn{2}{c}{\textbf{\textbf{\textbf{\textbf{\textbf{\textbf{\textbf{\textbf{Conversational Ability}}}}}}}}}  \\ 
\cmidrule(l){3-6}
                                 &                                   & \multicolumn{1}{l}{\textbf{MMLU $\uparrow$}} & \multicolumn{1}{l}{\textbf{Hellaswag $\uparrow$}}                   & \textbf{Vicuna $\uparrow$} & \textbf{MT-Bench $\uparrow$}                                                           \\ 
\midrule
\multirow{4}{*}{Mistral-7B-v0.1} & LoRA                              & \textbf{61.95}                               & \textbf{62.90}                                                      & 6.25                       & 4.68                                                                                   \\ 
\cmidrule(lr){2-6}
                                 & LoRA w/ ILA (10\%)               & 62.09                                        & 61.94                                                               & 5.99                       & 4.39                                                                                   \\
                                 & LoRA w/ ILA (20\%)                & 61.83                                        & 62.16                                                               & 6.12                       & 4.53                                                                                   \\
                                 & LoRA w/ ILA (30\%)                & 61.89                                        & 62.79                                                               & \textbf{6.27}              & \textbf{4.75}                                                                          \\
\bottomrule
\end{tabular}
}
\end{table}

\begin{table}[!t]
\renewcommand\arraystretch{0.8}
\caption{Comparison of QLoRA fine-tuning on \llama-7B vs. selectively fine-tuning important layers identified by \modelname{}. Evaluated on MMLU (5-shot), Hellaswag (0-shot), and GPT-4o scores for Vicuna and MT-Bench prompts (averaged over three runs). Grey cells indicate improvements over the base model by \modelname{}.} \label{table:qlora}
\centering
\resizebox{0.95\linewidth}{!}{
\begin{tabular}{llcccc} 
\toprule
\multirow{2}{*}{\textbf{Datasets}} & \multirow{2}{*}{\textbf{Methods}} & \multicolumn{2}{c}{\textbf{\textbf{\textbf{\textbf{Language Understanding}}}}}                   & \multicolumn{2}{c}{\textbf{\textbf{\textbf{\textbf{Conversational Ability}}}}}       \\ 
\cmidrule(lr){3-3}\cmidrule(lr){4-4}\cmidrule(lr){5-5}\cmidrule(lr){6-6}
                                   &                                   & \multicolumn{1}{l}{\textbf{MMLU $\uparrow$}} & \multicolumn{1}{l}{\textbf{Hellaswag $\uparrow$}} & \textbf{Vicuna $\uparrow$}               & \textbf{MT-Bench $\uparrow$}              \\ 
\midrule
\multirow{3}{*}{LIMA}              & QLoRA                             & 43.06                                        & 55.47                                             & 5.31                                     & 2.98                                      \\
                                   & QLoRA w/ ILA (75\%)               & {\cellcolor[rgb]{0.753,0.753,0.753}}43.48    & {\cellcolor[rgb]{0.753,0.753,0.753}}55.95         & {\cellcolor[rgb]{0.753,0.753,0.753}}5.56 & {\cellcolor[rgb]{0.753,0.753,0.753}}3.19  \\
                                   & QLoRA w/ ILA (30\%)               & {\cellcolor[rgb]{0.753,0.753,0.753}}44.01    & {\cellcolor[rgb]{0.753,0.753,0.753}}55.82         & 5.17                                     & {\cellcolor[rgb]{0.753,0.753,0.753}}3.01  \\
\bottomrule
\end{tabular}
}
\end{table}

These findings highlight the effectiveness of our layer selection strategy, optimizing resource use with minimal performance trade-offs. Our integration with QLoRA shows that fine-tuning a targeted subset of important layers improves both performance and memory efficiency during fine-tuning.

For a clearer understanding of GPU memory savings, we measured memory consumption for QLoRA, LoRA, Full Fine-Tuning, and versions fine-tuning only the key layers identified by \modelname{}. Results are presented in Table~\ref{table:gpu_memory} in Appendix~\ref{SubApp:GPU_Memory}.

\begin{table}[t]
\renewcommand\arraystretch{0.8}
\caption{Performance comparison of \modelname{}, random, and position-based layer selection for fine-tuning \llama-7B on the No Robots dataset. \textbf{RL1}/\textbf{RL2} freeze $K$ randomly selected layers (different seeds); \textbf{FL} and \textbf{LL} freeze the first and last $K$ layers, respectively. Blue highlights indicate lower performance than \modelname{}.}
\label{table:ablation_random}
\centering
\resizebox{0.8\linewidth}{!}{
\begin{tabular}{lcccc} 
\toprule
\multirow{2}{*}{\textbf{Methods}} & \multicolumn{2}{c}{\textbf{\textbf{\textbf{\textbf{\textbf{\textbf{\textbf{\textbf{Language Understanding}}}}}}}}} & \multicolumn{2}{c}{\textbf{\textbf{\textbf{\textbf{\textbf{\textbf{\textbf{\textbf{Conversational Ability}}}}}}}}}  \\ 
\cmidrule(l){2-5}
                                  & \multicolumn{1}{l}{\textbf{MMLU $\uparrow$}} & \multicolumn{1}{l}{\textbf{Hellaswag $\uparrow$}}                   & \textbf{Vicuna $\uparrow$} & \textbf{MT-Bench $\uparrow$}                                                           \\ 
\midrule
LoRA                              & 44.58                                        & 59.46                                                               & 5.78                       & 3.98                                                                                   \\
LoRA w/ RL 1                      & \textcolor{blue}{44.23}                      & 59.71                                                               & \textcolor{blue}{5.72}     & \textcolor{blue}{3.96}                                                                 \\
LoRA w/ RL 2                      & \textcolor{blue}{43.98}                      & \textcolor{blue}{59.11}                                             & \textcolor{blue}{5.62}     & \textcolor{blue}{3.89}                                                                 \\
LoRA w/ FL                        & \textcolor{blue}{44.02}                      & \textcolor{blue}{59.32}                                             & \textcolor{blue}{5.58}     & \textcolor{blue}{3.71}                                                                 \\
LoRA w/ LL                        & \textcolor{blue}{44.61}                      & \textcolor{blue}{59.21}                                             & \textcolor{blue}{5.65}     & \textcolor{blue}{3.99}                                                                 \\
LoRA w/ ILA                       & 45.78                                        & 59.65                                                               & 5.90                       & 4.15                                                                                   \\
\bottomrule
\end{tabular}
}
\end{table}

\subsection{Ablation Study}\label{sec:ablation}
\textbf{Observation 1: Our layer importance ranking algorithm is effective.}
We evaluated our algorithm by comparing it to a baseline that fine-tunes all layers and three alternatives: (1) \textbf{RL 1} and \textbf{RL 2}, which randomly freeze top-$K$ layers; (2) \textbf{FL}, freezing the first $K$ layers; and (3) \textbf{LL}, freezing the last $K$ layers. As shown in Table~\ref{table:ablation_random}, these naive strategies underperform. In contrast, our method effectively identifies and freezes the least critical layers, yielding notable gains in both efficiency and performance.

\textbf{Observation 2: The important scores calculated using LoRA are similar to those obtained through full fine-tuning.}
To assess whether LoRA-based approximations differ from full fine-tuning (FFT), we compared parameter updates from LoRA (i.e., Eq.~(\ref{eq:LoRA})) and FFT (i.e., $\Delta\boldsymbol{\parameter}_t = \boldsymbol{\parameter}_t - \boldsymbol{\parameter}_0$). For both methods, we derived layer importance scores and selected the top 75\% of layers, then calculated the Jaccard similarity between the layers. As shown in Table~\ref{table:lora_vs_FFT}, LoRA achieves nearly 83\% overlap with the important layers identified by FFT, reducing computational overhead while effectively ranking layer importance. The results show that LoRA provides a strong approximation of $\Delta\boldsymbol{\parameter}_t$ compared to $\boldsymbol{\parameter}_t - \boldsymbol{\parameter}_0$.

\begin{table}[!t]
\centering
\renewcommand\arraystretch{0.9}
\caption{Jaccard Similarity between important layers selected using Full Fine-Tuning and LoRA for \llama-7B. Top 75\% highest-scoring layers are determined as important layers.}\label{table:lora_vs_FFT}
\resizebox{0.8\linewidth}{!}{
\begin{tabular}{cccc} 
\toprule
\textbf{Datasets}           & \textbf{LIMA (FFT)} & \textbf{No Robots (FFT)} & \textbf{Alpaca-GPT4 (FFT)}  \\ 
\hline
\textbf{LIMA (LoRA)}        & 0.84                & 0.76                     & 0.83                        \\
\textbf{No Robots (LoRA)}   & 0.78                & 0.80                     & 0.81                        \\
\textbf{Alpaca-GPT4 (LoRA)} & 0.82                & 0.83                     & 0.86                        \\
\bottomrule
\end{tabular}
}
\end{table}

\textbf{Observation 3: Cross-dataset evaluation of layer importance enhances performance.}
Different datasets highlight subtle differences in important layers (Table~\ref{table:diff_dataset}). By intersecting the top-$K$ least important layers from LIMA, No Robots, and Alpaca-GPT4 and freezing them during fine-tuning (Table~\ref{table:cross}, Appendix~\ref{SubApp:Cross-dataset}), we found that cross-dataset evaluation yields better results than dataset-specific fine-tuning. This suggests that assessing layer importance across datasets leads to more robust fine-tuning.


\textbf{Observation 4: Cross-model transfer of layer importance rankings is feasible but less effective than cross-dataset transfer.}
Models sharing architecture but trained on different datasets show strong agreement in important layers (Jaccard similarity of 0.90 for the top 75\%, Table~\ref{table:diff_dataset}). This drops to 0.70 across architectures (Table~\ref{table:cross_model_similarity}, Appendix~\ref{SubApp:Cross-model}), indicating reduced transferability. Nonetheless, significant overlap suggests cross-architecture transfer remains viable. Fine-tuning Mistral-7B-v0.1 using rankings from \llama-7B on No Robots (Table~\ref{table:cross_model_exp}, Appendix~\ref{SubApp:Cross-model}) confirms that cross-model transfer can still perform well.


\textbf{Observation 5: \modelname{} is robust to the initialization of layer importance scores.} As demonstrated in Table~\ref{table:init_scores} (Appendix~\ref{SubApp:InitialScores}), our algorithm \modelname{} is resilient to varying initial importance scores ($s_0 = 4.0, 2.0, 1.0$), with minimal impact on final rankings. The stable Jaccard similarities for the top 75\% of layers during \llama-7B fine-tuning on LIMA confirm reliable convergence regardless of initialization.

\textbf{Observation 6: The computation cost of \modelname{} is low.}
\modelname{} runs in two stages: Stage 1 trains the model with LoRA until $\epsilon$-stability, and Stage 2 tunes importance weights ($\gamma_t$) with the backbone and LoRA frozen. For both \textsc{Llama~\small2}-7B and Mistral-7B-v0.1 (225 linear layers), Stage 1 takes 6671 ms per iteration, and Stage 2 takes 5343 ms. Stage 2 finishes in 11 minutes (128 batches). Most cost lies in Stage 1, but Table~\ref{table:diff_stage} shows only 25–50\% of training milestones are needed for strong performance.

\section{Conclusion and Discussion: Beyond LLM Alignment}
\begin{figure}[!ht]
\centering
\begin{minipage}[b]{0.46\linewidth}
    \centering
    \includegraphics[width=\linewidth]{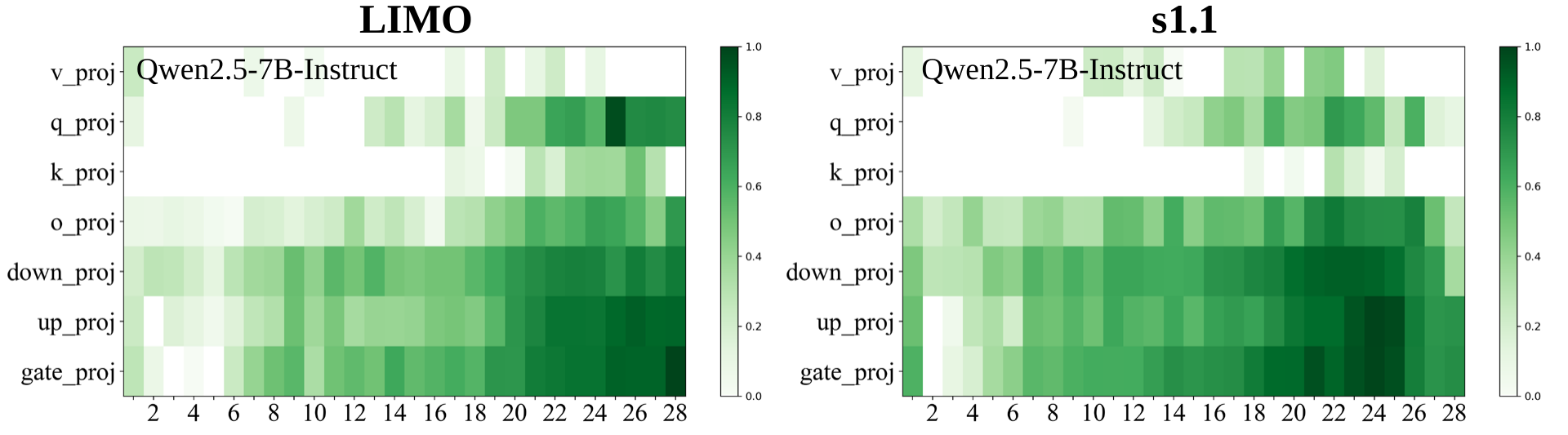}
    \caption{Layer-wise importance rankings for Qwen2.5-7B-Instruct fine-tuned using the LIMO and s1.1 datasets, respectively.}
    \label{fig:LIMO_vs_s1}
\end{minipage}
\hfill
\begin{minipage}[b]{0.5\linewidth}
    \centering
    \renewcommand\arraystretch{1.0}  
    \captionof{table}{Performance of Qwen2.5-7B-Instruct on mathematical reasoning benchmarks after fine-tuning with the LIMO dataset.}
    \label{table:LIMO_vs_s1}
    \resizebox{0.85\linewidth}{!}{
        \begin{tabular}{ccc} 
        \toprule
        \textbf{Methods} & \textbf{MATH500} & \textbf{AIME}  \\
        \midrule
        FFT             & 77.00            & 13.33         \\
        FFT w/ ILA      & \textbf{79.00}            & \textbf{16.67}         \\
        \bottomrule
        \end{tabular}
    }
    \vspace{1.8mm} 
\end{minipage}
\end{figure}

Our findings indicate that the alignment process of LLM imparts similar capabilities despite data variations. This complements prior research by revealing layer-specific roles and improving efficiency through strategic tuning and freezing of layers. Nonetheless, the approaches and insights derived from this study extend beyond LLM alignment.

\textbf{LLM Reasoning and Test-time Scaling.}
Advanced models like o1~\citep{openai2024learning}, Deepseek R1~\citep{guo2025deepseek}, and Kimi 1.5~\citep{team2025kimi} have exhibited strong reasoning capabilities. 
Similar to alignment, reasoning seeks to \emph{further activate the knowledge} acquired during pre-training. While alignment ensures that outputs align with human values, reasoning \emph{drives the model toward deeper inference for enhanced accuracy}. Rather than scaling model size or training data, recent studies such as LIMO~\citep{ye2025limo} and s1~\citep{muennighoff2025s1} investigate \textit{test-time scaling}—boosting performance by increasing the number of input tokens used for reasoning. Their findings show that even limited high-quality training data with chain-of-thought (CoT) examples can effectively enhance LLMs' reasoning capabilities.

To see whether our approaches can yield insights into LLM reasoning similar to those obtained in LLM alignment, we conducted preliminary experiments. We fine-tuned the Qwen2.5-7B-Instruct~\citep{yang2024qwen2} model using two different reasoning datasets, S1~\citep{muennighoff2025s1} and LIMO~\citep{ye2025limo}. Our observations reveal that the important layers identified from S1 and LIMO exhibit a Jaccard similarity of nearly \textbf{\textit{86\%}} (Fig.~\ref{fig:LIMO_vs_s1}), suggesting that the model acquires similar reasoning capabilities despite differences in datasets.
We also show that freezing the least important 25\% of layers enhances performance (Table~\ref{table:LIMO_vs_s1}), \textbf{highlighting the potential of selective fine-tuning for reasoning tasks}. For further details on the datasets and evaluation benchmarks, please refer to Appendix~\ref{App:LLM_reasoning}.


\bibliography{custom}
\bibliographystyle{colm2025_conference}

\appendix
\section{Proof of Theorem~\ref{theorem}}~\label{App:Proof}
The proof of Theorem~\ref{theorem} is presented as follows:
\begin{theorem}
For a sufficiently small $\epsilon$, $\parameter_T$ is $\epsilon$-stable, thus Assumption~\ref{assump:L-smooth} and Assumption~\ref{assump:bound} are satisfied. For any $t>T$, we assume that $\forall i, \gamma_t^i\in[0,1]$. Let $\rmgamma_t'$ denote the result of $\rmgamma_t$ after one step of gradient descent, i.e., $\rmgamma_t'=\rmgamma_t- \beta \nabla_{\rmgamma_t}\mathcal{L}(\parameter_t^{\rm mask})$. Then we have
\begin{equation} 
    \|\rmgamma_t'-\rmgamma_{t+1}'\|_2 \leq \beta (QL_2 + L_1)R\epsilon.
\end{equation}
\end{theorem}

\begin{proof}
Let $\hat{\rmgamma}$ be the initial values of $\rmgamma_t$ and $\rmgamma_{t+1}$. Then we have
\begin{equation} 
    \rmgamma_t' = \hat{\rmgamma} - \beta \nabla_{\rmgamma_t}\mathcal{L}(\parameter_t^{\rm mask}),
\end{equation}
\begin{equation} 
    \rmgamma_{t+1}' = \hat{\rmgamma} - \beta \nabla_{\rmgamma_{t+1}}\mathcal{L}(\parameter_{t+1}^{\rm mask}).
\end{equation}
The difference of $\rmgamma_t'$ and $\rmgamma_{t+1}'$ is 

\begin{align}
& \|\rmgamma_t' - \rmgamma_{t+1}'\|_2 
= \|(\hat{\rmgamma} - \beta \nabla_{\rmgamma_t}\mathcal{L}(\parameter_t^{\rm mask})) \notag\\
&\quad - (\hat{\rmgamma} - \beta \nabla_{\rmgamma_{t+1}}\mathcal{L}(\parameter_{t+1}^{\rm mask}))\|_2 \notag\\[6pt]
&= \beta \|\nabla_{\rmgamma_t}\mathcal{L}(\parameter_t^{\rm mask}) 
   - \nabla_{\rmgamma_{t+1}}\mathcal{L}(\parameter_{t+1}^{\rm mask})\|_2 \notag\\[6pt]
&= \beta\|\parameter_t \odot \nabla_{\parameter_t^{\rm mask}}\mathcal{L}(\parameter_{t}^{\rm mask})\notag\\[6pt]
&\quad - \parameter_{t+1} \odot \nabla_{\parameter_{t+1}^{\rm mask}}\mathcal{L}(\parameter_{t+1}^{\rm mask})\|_2 \notag\\[6pt]
&\leq \beta\|\parameter_t \odot (\nabla_{\parameter_t^{\rm mask}}\mathcal{L}(\parameter_t^{\rm mask})
   - \nabla_{\parameter_{t+1}^{\rm mask}}\mathcal{L}(\parameter_{t+1}^{\rm mask}))\|_2 \notag\\[6pt]
&\quad + \beta\|(\parameter_t - \parameter_{t+1}) \odot 
   \nabla_{\parameter_{t+1}^{\rm mask}}\mathcal{L}(\parameter_{t+1}^{\rm mask})\|_2.
\end{align}

Because $\mathcal{L}(\parameter)$ has an L-Lipschitz continuous gradient with constant $L_2>0$, and $\|\parameter_t\|\leq Q$, 

\noindent
\begin{align}
&\|\parameter_t \odot \nabla_{\parameter_{t}^{\rm mask}}\mathcal{L}(\parameter_{t}^{\rm mask})
- \parameter_{t+1}\odot \nabla_{\parameter_{t+1}^{\rm mask}}\mathcal{L}(\parameter_{t+1}^{\rm mask})\|_2\notag\\[6pt]
&\leq QL_2 \|\parameter_{t}^{\rm mask}-\parameter_{t+1}^{\rm mask}\|_2 \notag\\[6pt]
&= QL_2 \|\Delta\parameter_{t+1}-\Delta\parameter_t\|_2 \notag\\[6pt]
&= QL_2 \|\parameter_{t+1}-\parameter_t\|_2.
\end{align}
Because $\mathcal{L}(\parameter)$ is  $L$-smooth with constant $L_1$,

\noindent
\begin{align}
    \|(\parameter_t-\parameter_{t+1})\odot\nabla_{\parameter_{t+1}^{\rm mask}}\mathcal{L}(\parameter_{t+1}^{\rm mask})\|_2 \leq &
    L_1\|\parameter_t-\parameter_{t+1}\|.
\end{align}
Therefore,

\noindent
\begin{align}
    \|\rmgamma_t'-\rmgamma_{t+1}'\|_2 & \leq \beta (QL_2+L_1)\|\parameter_t-\parameter_{t+1}\|_2.
\end{align}
According to the Assumption~\ref{assump:bound}, we have $\|\parameter_t-\parameter_{t+1}\|_2\leq R\epsilon$, hence,
\begin{equation} 
    \|\rmgamma_t'-\rmgamma_{t+1}'\|_2 \leq \beta (QL_2 + L_1)R\epsilon.
\end{equation}
\end{proof}
\section{Experimental Setup}~\label{App:Exp_Setup}
\paragraph{\textbf{Datasets.}} (1) Alpaca-GPT4 contains 52K instruction-following data generated by GPT-4, utilizing prompts from Alpaca~\citep{alpaca}. (2) LIMA contains only 1K carefully curated prompts and responses. (3) No Robots contains 10K instructions and demonstrations created by skilled human annotators.

\paragraph{Models and Baselines.} We use four different models as the base for our experiments: \textsc{Llama~\small2}-7B~\citep{touvron2023llama}, \textsc{Llama~\small2}-13B, \textsc{Llama} 3.1-8B~\citep{dubey2024llama}, and Mistral-7B-v0.1~\citep{jiang2023mistral}. Our baselines are as follows: (1) \textbf{LoRA}\citep{hu2021lora}: Trainable rank decomposition matrices are added in parallel to existing weight matrices, including query/key/value projection ($W_q$, $W_k$, $W_v$), output projection ($W_o$) in self-attention, feed-forward networks ($W_{\rm up}$, $W_{\rm down}$, $W_{\rm gate}$), and the output layer ($W_{\mathrm{head}}$). (2) \textbf{AdaLoRA}\citep{zhang2023adaptive}: It dynamically adjusts the rank of incremental matrices to control the parameter budget, with AdaLoRA modules added to all linear layers, similar to LoRA. (3) \textbf{Full Fine-tune}: All model parameters, initialized from pre-trained weights and biases, undergo gradient updates during fine-tuning.

\paragraph{Evaluation and Training Setup.} We assess language model alignment across two key dimensions:
(1) \textbf{Language Understanding Ability}: Evaluated using \textbf{MMLU}~\citep{hendrycks2021measuring} for specialized knowledge and \textbf{Hellaswag}~\citep{zellers2019hellaswag} for commonsense reasoning.
(2) \textbf{Conversational Ability}: Measured using \textbf{MT-Bench}~\citep{zheng2023judging} (multi-turn) and \textbf{Vicuna}~\citep{chiang2023vicuna} (single-turn), with responses graded by \textbf{\mbox{GPT-4o}}. All evaluations are performed three times, and the average scores are reported. \textbf{We conduct hyperparameter searches for LoRA and full fine-tuning to establish strong baselines.}

\paragraph{Targeted Performance.} (1) \textbf{Language Understanding Ability}: Recent research~\citep{du2020self,sun2021ernie,dubey2024llama} suggests that the learning of language understanding tasks essentially occurs during the pre-training phase of the base model. Therefore, significant performance improvements in language understanding tasks (i.e., MMLU, Hellaswag) after alignment are not expected. However, \emph{it is crucial to ensure the model retains the learned knowledge during alignment}. 
(2) \textbf{Conversational Ability}: Without alignment, the pre-train model's conversational ability is poor. For example, \textsc{Llama~\small2}-7B often produces incorrect or irrelevant responses on the Vicuna dataset. However, \emph{its conversational ability can be significantly improved through the alignment process}.

For all experiments, we follow fine-tuning hyperparameters: we use AdamW with $\beta_1$ = 0.9, $\beta_2$ = 0.99 and weight decay of $0.1$. The scheduler employed is a cosine scheduler with a warmup ratio of 0.01. For LoRA baselines, we set the hyperparameter rank $r$ as 32.

\subsection{No Robots Dataset}
We do a hyperparameter search for LoRA over the following variables: learning rate $\{0.001, 0.002, 0.0005, 0.0002, 0.0001\}$, training epochs $\{2,3,4,5\}$. We do hyperparameter search for full fine-tuning over the following variables: learning rate $\{1e-4, 2e-5, 1e-5, 5e-6, 2e-6\}$, training epochs $\{2,3,4,5\}$.

\textbf{\textsc{Llama~\small2}-7B}. 
Both LoRA and AdaLoRA use a dropout rate of $0.1$ and a learning rate of $0.001$. The number of training epochs is $3$. For full fine-tuning, the learning rate is set to $0.00001$, with the number of training epochs also being $3$. The training parameters for ILA are consistent with those of the baselines.

\textbf{Mistral-7B.} For LoRA and AdaLorA, we set the dropout rate as $0.1$. The learning is $0.0002$. The number of training epochs is $2$. For full fine-tuning, the learning rate is set as $0.000002$ and the number of training epochs is $2$. The training parameters of ILA are the same as the baselines. 

\subsection{LIMA Dataset}
We do a hyperparameter search for LoRA over the following variables: learning rate $\{0.001, 0.002, 0.0005, 0.0002, 0.0001\}$, training epochs $\{5, 10, 15, 20\}$. We do hyperparameter search for full fine-tuning over the following variables: learning rate $\{1e-4, 2e-5, 1e-5, 5e-6, 2e-6\}$, training epochs $\{5, 10, 15, 20\}$.

\paragraph{\textsc{Llama~\small2}-7B}. For LoRA and AdaLorA, we set the dropout rate as $0.1$. The learning is $0.001$. The number of training epochs is $20$. For full fine-tuning, the learning rate is set as $0.00001$ and the number of training epochs is $5$. The training parameters of ILA are the same as the baselines.

\paragraph{Mistral-7B.} For LoRA and AdaLorA, we set the dropout rate as $0.1$. The learning is $0.0002$. The number of training epochs is $5$. For full fine-tuning, the learning rate is set as $0.000005$ and the number of training epochs is $5$. The training parameters of ILA are the same as the baselines.

\subsection{Alpaca-GPT Dataset.}
We do a hyperparameter search for LoRA over the following variables: learning rate $\{0.001, 0.002, 0.0005, 0.0002, 0.0001\}$, training epochs $\{0.5, 1, 1.5, 2, 3\}$. We do hyperparameter search for full fine-tuning over the following variables: learning rate $\{1e-4, 2e-5, 1e-5, 5e-6, 2e-6\}$, training epochs $\{0.5, 1, 1.5, 2, 3\}$.

\paragraph{\textsc{Llama~\small2}-7B}. For LoRA and AdaLorA, we set the dropout rate as $0.1$. The learning is $0.0002$. The number of training epochs is $1.5$. For full fine-tuning, the learning rate is set as $0.000002$ and the number of training epochs is $0.5$. The training parameters of ILA are the same as the baselines.

\paragraph{Mistral-7B.} For LoRA and AdaLorA, we set the dropout rate as $0.1$. The learning is $0.0002$. The number of training epochs is $5$. For full fine-tuning, the learning rate is set as $0.000002$ and the number of training epochs is $0.5$. The training parameters of ILA are the same as the baselines.
\section{Additional Experiments}~\label{App:Extra_Exp}
\subsection{GPU Memory Usage Analysis}~\label{SubApp:GPU_Memory}
We provide an overview of the GPU memory usage for different fine-tuning strategies, as shown in Table~\ref{table:gpu_memory}. The table demonstrates the GPU memory usage and average training time per iteration for various fine-tuning approaches, including LoRA, QLoRA, full fine-tune, and their modified versions where only 30\% of the important layers identified by \modelname{} are fine-tuned. Both LoRA and QLoRA show substantial reductions in memory usage when restricted to tuning only 30\% of important layers, compared to the full-layer fine-tuning approaches. These results indicate that selectively fine-tuning a small set of critical layers is highly effective in reducing GPU memory consumption, particularly for efficient methods like QLoRA. This suggests that targeted fine-tuning can enhance computational efficiency while preserving model performance, which is especially beneficial when scaling large language models with limited hardware resources.

\begin{table}[!ht]
\renewcommand\arraystretch{1.0}
\caption{GPU memory usage for LoRA, QLoRA, Full Fine-tune and LoRA/QLoRA/Full Fine-tune with only 30\% of important layers fine-tuned. Batch size is set to 2, and the maximum token length is 1024. Percentages in parentheses indicate the proportion of linear layers fine-tuned.} \label{table:gpu_memory}
\centering
\resizebox{0.8\linewidth}{!}{
\begin{tabular}{lcc} 
\toprule
                            & \begin{tabular}[c]{@{}c@{}}\textbf{GPU}\\\textbf{Memory Usage (MiB)}\end{tabular} & \multicolumn{1}{l}{\textcolor[rgb]{0.2,0.2,0.2}{\textbf{Training time (ms)}}}  \\ 
\midrule
Full Fine-tune (100\%)       & 81276                                                                             & 396                                                                            \\
Full Fine-tune w/ ILA (30\%) & 33458                                                                             & 304                                                                            \\
LoRA (100\%)                & 32752                                                                             & 403                                                                            \\
LoRA w/ ILA (30\%)          & 28586                                                                             & 359                                                                            \\
QLoRA~(100\%)               & 26238                                                                             & 523                                                                            \\
QLoRA w/ ILA (30\%)         & 17912                                                                             & 423                                                                            \\
\bottomrule
\end{tabular}
}
\end{table}

\begin{table*}[!ht]
\renewcommand\arraystretch{1.0}
\caption{Results of fine-tuning Mistral-7B-v0.1 on the LIMA dataset using \modelname{} to identify important layers from various datasets. \textbf{Dataset (Imp. Layers)} indicates the datasets utilized to search for the important layers. \textbf{Intersection} represents freezing the layers that are the intersection of the top-$K$ least important layers found from the LIMA, No Robots, and Alpaca GPT4 datasets. Evaluated using MMLU (5-shot), Hellaswag (0-shot), \textbf{GPT-4 scores} on Vicuna prompts, and
MT-Bench prompts.} \label{table:cross}
\centering
\resizebox{0.7\linewidth}{!}{
\begin{tabular}{cccccc} 
\toprule
\multirow{2}{*}{\begin{tabular}[c]{@{}c@{}}\textbf{Dataset }\\\textbf{ (Imp. Layers)}\end{tabular}} & \multirow{2}{*}{\begin{tabular}[c]{@{}c@{}}\textbf{Dataset }\\\textbf{ (Fine-tune)}\end{tabular}} & \multicolumn{2}{c}{\textbf{\textbf{\textbf{\textbf{\textbf{\textbf{\textbf{\textbf{\textbf{\textbf{\textbf{\textbf{\textbf{\textbf{\textbf{\textbf{Language Understanding}}}}}}}}}}}}}}}}} & \multicolumn{2}{c}{\textbf{\textbf{\textbf{\textbf{\textbf{\textbf{\textbf{\textbf{\textbf{\textbf{\textbf{\textbf{\textbf{\textbf{\textbf{\textbf{Conversational Ability}}}}}}}}}}}}}}}}}  \\ 
\cmidrule(lr){3-4}\cmidrule(lr){5-6}
                                                                                                    &                                                                                                  & \multicolumn{1}{l}{\textbf{MMLU $\uparrow$}} & \multicolumn{1}{l}{\textbf{Hellaswag $\uparrow$}}                                                                                           & \textbf{Vicuna $\uparrow$} & \textbf{MT-Bench $\uparrow$}                                                                                                                                   \\ 
\midrule
LIMA                                                                                                & LIMA                                                                                             & \textbf{61.82}                               & 65.48                                                                                                                                       & 6.99                       & 5.38                                                                                                                                                           \\
No Robots                                                                                           & LIMA                                                                                             & 61.52                                        & 65.51                                                                                                                                       & 6.92                       & 5.34                                                                                                                                                           \\
Alpaca-GPT4                                                                                         & LIMA                                                                                             & 61.23                                        & 65.20                                                                                                                                       & 7.03                       & 5.21                                                                                                                                                           \\
Intersection                                                                                        & LIMA                                                                                             & 61.49                                        & \textbf{65.62}                                                                                                                              & \textbf{7.06}              & \textbf{5.44}                                                                                                                                                  \\
\bottomrule
\end{tabular}
}
\end{table*}
\subsection{Cross-dataset Evaluation of Layer Importance}~\label{SubApp:Cross-dataset} 
As shown in Table~\ref{table:diff_dataset}, different datasets reveal subtle variations in the layers identified as important. This suggests that layers consistently deemed unimportant across multiple datasets are likely genuinely non-essential. To validate this, we intersect the top-$K$ least important layers identified from three datasets (LIMA, No Robots, and Alpaca-GPT4) to derive a set of universally non-critical layers. The results are presented in Table~\ref{table:cross}.

Our analysis reveals that a holistic consideration of layer importance across multiple datasets yields superior results compared to dataset-specific approaches. For instance, identifying important layers within the LIMA dataset and fine-tuning on the No Robots dataset is less effective than an integrated approach. Similarly, finding important layers and fine-tuning exclusively on the No Robots dataset do not perform as well as the comprehensive method. This suggests that a cross-dataset evaluation of layer importance can lead to more robust and effective fine-tuning strategies.

\subsection{Cross-model Transfer of Layer Importance Rankings}~\label{SubApp:Cross-model}
We evaluated the transferability of layer importance rankings across different models, focusing on the Jaccard similarity of the top 75\% important layers. The results for various models are shown in Table~\ref{table:cross_model_similarity}. As seen, the Jaccard similarity between different architectures is approximately \textbf{0.70}, suggesting a moderate overlap in the layers identified as important.

\begin{table}[h!]
\centering
\caption{Jaccard similarities of the top 75\% important layers across different models.}\label{table:cross_model_similarity}
\resizebox{0.6\linewidth}{!}{
\begin{tabular}{cccc} 
\toprule
                                                                                          & \begin{tabular}[c]{@{}c@{}}\textbf{\textbf{\llama}-7B }\\\textbf{(LIMA)}\end{tabular} & \begin{tabular}[c]{@{}c@{}}\textbf{\textbf{\llama}-7B }\\\textbf{(NoRobots)}\end{tabular} & \begin{tabular}[c]{@{}c@{}}\textbf{\llama-7B }\\\textbf{(Alpaca-GPT4)}\end{tabular}  \\ 
\hline
\begin{tabular}[c]{@{}c@{}}\textbf{Mistral-7B-v0.1 }\\\textbf{(LIMA)}\end{tabular}        & 0.67                                                                                                 & -                                                                                                        & -                                                                                                   \\
\begin{tabular}[c]{@{}c@{}}\textbf{Mistral-7B-v0.1 }\\\textbf{(NoRobots)}\end{tabular}    & 0.70                                                                                                 & 0.71                                                                                                     & -                                                                                                   \\
\begin{tabular}[c]{@{}c@{}}\textbf{Mistral-7B-v0.1 }\\\textbf{(Alpaca-GPT4)}\end{tabular} & 0.71                                                                                                 & 0.66                                                                                                     & 0.75                                                                                                \\
\bottomrule
\end{tabular}
}

\end{table}

In our subsequent experiment, we used \llama-7B's layer importance rankings from the No Robots dataset to fine-tune Mistral-7B-v0.1, as shown in Table~\ref{table:cross_model_exp}. The results demonstrate that while cross-model transfer offers performance improvements, fine-tuning using rankings from the same model architecture yields the best results.

These findings suggest that cross-model transfer of layer importance rankings is possible, though less effective than using rankings from the same architecture. Fine-tuning the top 75\% of layers based on cross-model transfer shows some improvement, while fine-tuning only the top 30\% achieves comparable performance.

\begin{table*}[!ht]
\centering
\caption{Experimental results for Mistral-7B-v0.1 on the No Robots dataset, using layer importance rankings derived from \llama-7B.}\label{table:cross_model_exp}
\resizebox{1.0\linewidth}{!}{
\begin{tabular}{ccccc} 
\toprule
\textbf{Methods}                                   & \textbf{MMLU}  & \textbf{Hellaswag} & \textbf{Vicuna} & \textbf{MT-Bench}  \\ 
\hline
\textbf{LoRA}                                      & 61.95          & 62.90              & 6.25            & 4.68               \\ 
\hline
\textbf{LoRA w/ ILA (75\%) (cross-model transfer)} & 62.10          & \textbf{63.21}     & 6.29            & 4.72               \\
\textbf{LoRA w/ ILA (75\%)}                        & \textbf{62.14} & 62.80              & \textbf{6.42}   & \textbf{4.87}      \\ 
\hline
\textbf{LoRA w/ ILA (30\%) (cross-model transfer)} & 61.77          & \textbf{63.16}     & 6.11            & 4.60               \\
\textbf{LoRA w/ ILA (30\%)}                        & \textbf{61.89} & 62.79              & \textbf{6.27}   & \textbf{4.75}      \\
\bottomrule
\end{tabular}
}
\end{table*}

\subsection{Impact of Initial Layer Importance Scores}~\label{SubApp:InitialScores}
We evaluated the effect of different initial layer importance scores on the consistency of identified important layers. The scores were initialized to $s_0 = 4.0, 2.0, 1.0$. The consistency was measured using the Jaccard similarity of the top 75\% important layers identified during fine-tuning of LLama 2-7B on the LIMA dataset. The results are shown in Table~\ref{table:init_scores}. These results show that the method is stable with respect to initialization. While the choice of $s_t$ can influence the optimization trajectory, it does not significantly impact the convergence or final importance rankings.

\begin{table}[h!]
\centering
\caption{The Jaccard similarities of top 75\% important layers identified during fine-tuning of \llama-7B on the LIMA dataset with varying initial scores.} \label{table:init_scores}
\resizebox{0.45\linewidth}{!}{
\begin{tabular}{cccc} 
\toprule
\textbf{Initial Scores} & \textbf{4.0} & \textbf{2.0} & \textbf{1.0}  \\ 
\hline
\textbf{4.0}            & -            & -            & -             \\
\textbf{2.0}            & 0.83         & -            & -             \\
\textbf{1.0}            & 0.78         & 0.88         & -             \\
\bottomrule
\end{tabular}
}
\end{table}

\subsection{Additional Experiments on Model Scalability}~\label{SubApp:Extra_Exp}
To assess whether freezing unimportant layers continues to enhance model performance at a larger scale, we conducted additional experiments on \llama-13B. Specifically, we fine-tuned \llama-13B using the No Robots and LIMA datasets, with results compared against LoRA presented in the table below. The experimental outcomes demonstrate that our method maintains strong performance on \llama-13B. Despite the increased model size, the underlying architectural similarities suggest that our approach remains effective and scalable, likely extending its benefits to even larger models.

We also carried out further experiments on \textbf{LIMA} and \textbf{Alpaca-GPT4} using \textbf{\llama-7B}, \textbf{Mistral-7B-v0.1} and \textbf{Llama 3.1-8B} to evaluate the adaptability of our approach across different model architectures. Consistently, our method outperformed LoRA while requiring fewer layers to be fine-tuned. These findings further validate the robustness and scalability of our approach, showing its capability to effectively enhance performance across various model sizes and architectural variations.

\begin{table*}[!ht]
\renewcommand\arraystretch{1.0}
\caption{Fine-tuning results of \llama-13B on the LIMA and No Robots datasets. Evaluated using MMLU (5-shot), Hellaswag (0-shot), \textbf{GPT-4o scores} on Vicuna prompts, and
MT-Bench prompts. Cells highlighted in grey indicate that \modelname{} has improved the performance of the base model.} \label{table:llama13B}
\centering
\resizebox{0.7\linewidth}{!}{
\begin{tabular}{llcccc} 
\toprule
\multirow{2}{*}{\textbf{Datasets}} & \multirow{2}{*}{\textbf{Methods}} & \multicolumn{2}{c}{\textbf{\textbf{\textbf{\textbf{Language Understanding}}}}}                   & \multicolumn{2}{c}{\textbf{\textbf{\textbf{\textbf{Conversational Ability}}}}}       \\ 
\cmidrule(l){3-6}
                                   &                                   & \multicolumn{1}{l}{\textbf{MMLU $\uparrow$}} & \multicolumn{1}{l}{\textbf{Hellaswag $\uparrow$}} & \textbf{Vicuna $\uparrow$}               & \textbf{MT-Bench $\uparrow$}              \\ 
\midrule
\multirow{2}{*}{LIMA}              & LoRA                              & 53.85                                        & 63.08                                             & 6.16                                     & 3.79                                      \\
                                   & LoRA w/ ILA                       & {\cellcolor[rgb]{0.753,0.753,0.753}}54.33    & 62.04                                             & {\cellcolor[rgb]{0.753,0.753,0.753}}6.25 & {\cellcolor[rgb]{0.753,0.753,0.753}}3.91  \\ 
\midrule
\multirow{2}{*}{No Robots}         & LoRA                              & 54.08                                        & 61.73                                             & 5.72                                     & 4.24                                      \\
                                   & LoRA w/ ILA                       & {\cellcolor[rgb]{0.753,0.753,0.753}}54.45    & 61.13                                             & {\cellcolor[rgb]{0.753,0.753,0.753}}5.88 & {\cellcolor[rgb]{0.753,0.753,0.753}}4.37  \\
\bottomrule
\end{tabular}
}
\end{table*}

\begin{table*}[!t]
\renewcommand\arraystretch{1.1}
\caption{Comparative evaluation of \textsc{Llama~\small2}-7B, Mistral-7B-v0.1, and Llama 3.1-8B models fine-tuned on the LIMA Dataset. 
Evaluated using MMLU (5-shot), Hellaswag (0-shot), \textbf{GPT-4o scores} on Vicuna prompts, and MT-Bench prompts. \textbf{The evaluations are performed three times, and the average scores are reported.}
Cells highlighted in grey indicate that \modelname{} has enhanced the performance of the base model. The best result is marked in bold.} \label{table:lima_all}
\centering
\resizebox{0.8\linewidth}{!}{
\begin{tabular}{llcccc} 
\toprule
\multirow{2}{*}{\textbf{Models}} & \multirow{2}{*}{\textbf{Methods}} & \multicolumn{2}{c}{\textbf{\textbf{Language Understanding}}}                                            & \multicolumn{2}{c}{\textbf{\textbf{Conversational Ability}}}                                                             \\ 
\cmidrule(l){3-6}
                                 &                                   & \multicolumn{1}{l}{\textbf{MMLU $\uparrow$}}       & \multicolumn{1}{l}{\textbf{Hellaswag $\uparrow$}}  & \textbf{Vicuna $\uparrow$}                                 & \textbf{MT-Bench $\uparrow$}                                \\ 
\midrule
\multirow{5}{*}{\llama-7B}        & AdaLoRA                           & 44.21                                              & 59.85                                              & 5.22                                                       & 3.51                                                        \\ 
\cmidrule(lr){2-6}
                                 & Full Fine-tune                     & \textbf{46.36}                                     & 62.06                                              & 5.83                                                       & 3.71                                                        \\
                                 & Full Fine-tune w/ ILA              & 46.32                                              & {\cellcolor[rgb]{0.753,0.753,0.753}}\textbf{62.18} & {\cellcolor[rgb]{0.753,0.753,0.753}}\textbf{\textbf{5.98}} & {\cellcolor[rgb]{0.753,0.753,0.753}}\textbf{\textbf{3.85}}  \\ 
\cmidrule(lr){2-6}
                                 & LoRA                              & 43.18                                              & 54.52                                              & 5.43                                                       & 3.45                                                        \\
                                 & LoRA w/ ILA                       & {\cellcolor[rgb]{0.753,0.753,0.753}}44.13          & {\cellcolor[rgb]{0.753,0.753,0.753}}54.55          & {\cellcolor[rgb]{0.753,0.753,0.753}}5.62                   & {\cellcolor[rgb]{0.753,0.753,0.753}}3.72                    \\ 
\midrule
\multirow{5}{*}{Mistral-7B-v0.1} & AdaLoRA                           & \textbf{62.40}                                     & 61.52                                              & 6.64                                                       & 4.49                                                        \\ 
\cmidrule(lr){2-6}
                                 & Full Fine-tune                     & 60.11                                              & 63.76                                              & 6.88                                                       & 4.63                                                        \\
                                 & Full Fine-tune w/~ILA              & {\cellcolor[rgb]{0.753,0.753,0.753}}61.01          & {\cellcolor[rgb]{0.753,0.753,0.753}}64.01          & {\cellcolor[rgb]{0.753,0.753,0.753}}6.95                   & {\cellcolor[rgb]{0.753,0.753,0.753}}\textbf{4.77}           \\ 
\cmidrule(lr){2-6}
                                 & LoRA                              & 60.83                                              & 65.42                                              & 6.70                                                       & 4.58                                                        \\
                                 & LoRA w/~ILA                       & {\cellcolor[rgb]{0.753,0.753,0.753}}61.52          & {\cellcolor[rgb]{0.753,0.753,0.753}}\textbf{65.51} & {\cellcolor[rgb]{0.753,0.753,0.753}}\textbf{6.98}          & {\cellcolor[rgb]{0.753,0.753,0.753}}4.69                    \\ 
\hline
\multirow{5}{*}{Llama 3.1-8B}    & AdaLoRA                           & 63.55                                              & 62.65                                              & 6.50                                                       & 4.73                                                        \\ 
\cline{2-6}
                                 & Full Fine-tune                     & 64.31                                              & 65.64                                              & 7.09                                                       & 5.12                                                        \\
                                 & Full Fine-tune w/~ILA              & {\cellcolor[rgb]{0.753,0.753,0.753}}\textbf{64.73} & {\cellcolor[rgb]{0.753,0.753,0.753}}\textbf{65.98} & {\cellcolor[rgb]{0.753,0.753,0.753}}\textbf{7.17}          & {\cellcolor[rgb]{0.753,0.753,0.753}}\textbf{5.23}           \\ 
\cline{2-6}
                                 & LoRA                              & 62.33                                              & 62.92                                              & 6.57                                                       & 4.79                                                        \\
                                 & LoRA w/~ILA                       & {\cellcolor[rgb]{0.753,0.753,0.753}}63.31          & {\cellcolor[rgb]{0.753,0.753,0.753}}63.01          & {\cellcolor[rgb]{0.753,0.753,0.753}}6.61                   & {\cellcolor[rgb]{0.753,0.753,0.753}}4.93                    \\
\bottomrule
\end{tabular}
}
\end{table*}

\begin{table*}[t]
\renewcommand\arraystretch{1.0}
\caption{Comparative evaluation of \textsc{Llama~\small2}-7B, Mistral-7B-v0.1, and Llama 3.1-8B models fine-tuned on the Alpaca-GPT4 Dataset. 
Evaluated using MMLU (5-shot), Hellaswag (0-shot), \textbf{GPT-4o scores} on Vicuna prompts, and MT-Bench prompts. \textbf{The evaluations are performed three times, and the average scores are reported.}
Cells highlighted in grey indicate that \modelname{} has enhanced the performance of the base model. The best result is marked in bold.} \label{table:alpaca_all}
\centering
\resizebox{0.8\linewidth}{!}{
\begin{tabular}{llcccc} 
\toprule
\multirow{2}{*}{\textbf{Models}} & \multirow{2}{*}{\textbf{Methods}} & \multicolumn{2}{c}{\textbf{\textbf{\textbf{\textbf{Language Understanding}}}}}                          & \multicolumn{2}{c}{\textbf{\textbf{\textbf{\textbf{Conversational Ability}}}}}                         \\ 
\cmidrule(l){3-6}
                                 &                                   & \multicolumn{1}{l}{\textbf{MMLU $\uparrow$}}       & \multicolumn{1}{l}{\textbf{Hellaswag $\uparrow$}}  & \textbf{Vicuna $\uparrow$}                        & \textbf{MT-Bench $\uparrow$}                       \\ 
\midrule
\multirow{5}{*}{llama-7B}        & AdaLoRA                           & 46.13                                              & 57.85                                              & 6.89                                              & 3.78                                               \\ 
\cmidrule(lr){2-6}
                                 & Full Fine-tune                     & 45.91                                              & 57.73                                              & 6.78                                              & 3.72                                               \\
                                 & Full Fine-tune w/ ILA              & {\cellcolor[rgb]{0.753,0.753,0.753}}\textbf{46.23} & 57.67                                              & {\cellcolor[rgb]{0.753,0.753,0.753}}6.99          & {\cellcolor[rgb]{0.753,0.753,0.753}}3.85           \\ 
\cmidrule(lr){2-6}
                                 & LoRA                              & 43.66                                              & \textbf{58.49}                                     & 6.96                                              & 3.80                                               \\
                                 & LoRA w/ ILA                       & {\cellcolor[rgb]{0.753,0.753,0.753}}44.69          & 58.22                                              & {\cellcolor[rgb]{0.753,0.753,0.753}}7.17          & {\cellcolor[rgb]{0.753,0.753,0.753}}3.99           \\ 
\midrule
\multirow{5}{*}{Mistral-7B-v0.1} & AdaLoRA                           & \textbf{62.48}                                     & 62.08                                              & 7.25                                              & 4.77                                               \\ 
\cmidrule(lr){2-6}
                                 & Full Fine-tune                     & 60.56                                              & 62.80                                              & 7.19                                              & 4.78                                               \\
                                 & Full Fine-tune w/ ILA              & {\cellcolor[rgb]{0.753,0.753,0.753}}60.88          & {\cellcolor[rgb]{0.753,0.753,0.753}}\textbf{62.91} & {\cellcolor[rgb]{0.753,0.753,0.753}}7.35          & {\cellcolor[rgb]{0.753,0.753,0.753}}4.91           \\ 
\cmidrule(lr){2-6}
                                 & LoRA                              & 61.82                                              & 62.70                                              & 7.23                                              & 4.89                                               \\
                                 & LoRA w/ ILA                       & {\cellcolor[rgb]{0.753,0.753,0.753}}62.14          & {\cellcolor[rgb]{0.753,0.753,0.753}}62.80          & {\cellcolor[rgb]{0.753,0.753,0.753}}7.33          & {\cellcolor[rgb]{0.753,0.753,0.753}}5.02           \\ 
\hline
\multirow{5}{*}{Llama 3.1-8B}    & AdaLoRA                           & \textbf{65.82}                                     & 61.02                                              & 7.48                                              & 5.39                                               \\ 
\cline{2-6}
                                 & Full Fine-tune                     & 63.58                                              & 61.58                                              & 7.33                                              & 5.32                                               \\
                                 & Full Fine-tune w/ ILA              & {\cellcolor[rgb]{0.753,0.753,0.753}}64.61          & {\cellcolor[rgb]{0.753,0.753,0.753}}61.74          & {\cellcolor[rgb]{0.753,0.753,0.753}}7.57          & {\cellcolor[rgb]{0.753,0.753,0.753}}5.42           \\ 
\cline{2-6}
                                 & LoRA                              & 65.40                                              & 61.72                                              & 7.65                                              & 5.43                                               \\
                                 & LoRA w/ ILA                       & {\cellcolor[rgb]{0.753,0.753,0.753}}65.76          & {\cellcolor[rgb]{0.753,0.753,0.753}}\textbf{61.81} & {\cellcolor[rgb]{0.753,0.753,0.753}}\textbf{7.79} & {\cellcolor[rgb]{0.753,0.753,0.753}}\textbf{5.55}  \\
\bottomrule
\end{tabular}
}
\end{table*}
\section{ILA for LLM Reasoning}~\label{App:LLM_reasoning}
\paragraph{Datasets.} (1) \textbf{LIMO~\citep{ye2025limo}:} This dataset comprises 817 carefully selected problems drawn from an initial pool of tens of millions. The final selection meets strict quality standards and covers a broad range of mathematical reasoning tasks. High-quality solutions are provided by both human experts and AI systems like DeepSeek R1~\citep{guo2025deepseek}.
(2) \textbf{s1.1~\citep{muennighoff2025s1}:} This dataset includes 1,000 questions paired with reasoning traces, curated based on three rigorously validated criteria: difficulty, diversity, and quality. The chain-of-thought solutions are generated by DeepSeek R1~\citep{guo2025deepseek}.

\paragraph{Evaluation.} (1) AIME24: This set contains 30 problems from the 2024 American Invitational Mathematics Examination (AIME), administered on January 31 and February 1, 2024.
(2) MATH500~\citep{hendrycks2021measuring}: A benchmark consisting of competition-level math problems spanning a range of difficulties.

\paragraph{Consistency in Layer Importance Across Datasets.} 
We applied our proposed ILA algorithm to identify layer importance rankings on both the LIMO and s1.1 datasets using the Qwen2.5-7B-Instruct model~\citep{yang2024qwen2}. The resulting rankings showed strong consistency, with a Jaccard similarity of \textbf{\textit{0.86}} (see Fig.\ref{fig:LIMO_vs_s1}), suggesting that LLMs tend to acquire similar reasoning-related knowledge across datasets. We hypothesize that much of this knowledge is already learned during pretraining—as also suggested by LIMO~\citep{ye2025limo}—and that fine-tuning primarily serves to activate the model’s latent reasoning abilities.

Based on the identified importance rankings, we further conducted experiments on LIMO by freezing approximately 25\% of the least important layers. As shown in Table~\ref{table:LIMO_vs_s1}, fine-tuning only the top 75\% most important layers led to a slight improvement in performance, indicating that selective tuning can help enhance the model’s reasoning capabilities.

\end{document}